\documentclass[10pt, a4paper]{article}
\usepackage{lrec2022} 
\usepackage{multibib}
\newcites{languageresource}{Language Resources}
\usepackage{graphicx}
\usepackage{tabularx}
\usepackage{soul}
\usepackage{titlesec}
\titleformat{\section}{\normalfont\large\bfseries\center}{\thesection.}{1em}{}
\titleformat{\subsection}{\normalfont\SmallTitleFont\bfseries\raggedright}{\thesubsection.}{1em}{}
\titleformat{\subsubsection}{\normalfont\normalsize\bfseries\raggedright}{\thesubsubsection.}{1em}{}
\renewcommand\thesection{\arabic{section}}
\renewcommand\thesubsection{\thesection.\arabic{subsection}}
\renewcommand\thesubsubsection{\thesubsection.\arabic{subsubsection}}

\usepackage{epstopdf}
\usepackage[utf8]{inputenc}

\usepackage{hyperref}
\hypersetup{
    colorlinks = true,
    allcolors = {blue}
}
\usepackage{xstring}

\usepackage{color}

\usepackage[acronym]{glossaries}

\makeglossaries

\newacronym{nlp}{NLP}{Natural Language Processing}
\newacronym{ner}{NER}{Named Entity Recognition}
\newacronym{sa}{SA}{Sentiment Analysis}
\newacronym{ml}{ML}{Machine Learning}
\newacronym{bow}{BoW}{bag-of-words}
\newacronym{cbow}{CBoW}{continuous Bag-of-Words}
\newacronym{sltc}{SLTC}{Swedish Language Technology Conference}
\newacronym{ann}{ANN}{artificial neural network}
\newacronym{nn}{NN}{neural network}
\newacronym{lstm}{LSTM}{Long Short Term Memory Network}
\newacronym{bilstm}{biLSTM}{bidirectional Long Short Term Memory Network}
\newacronym{sota}{SoTA}{state-of-the-art}
\newacronym{nlg}{NLG}{Natural Language Generation}
\newacronym{nlu}{NLU}{Natural Language Understanding}
\newacronym{mwe}{MWE}{Multi-Word Expression}
\newacronym{sw}{SW}{Simple Wiki}
\newacronym{mt}{MT}{Machine Translation}
\newacronym{bw}{BW}{Billion Word}
\newacronym{pie}{PIE}{Potential Idiomatic Expression}
\newacronym{iaa}{IAA}{Inter-Annotator Agreement}
\newacronym{rte}{RTE}{Recognizing Textual Entailment}
\newacronym{ir}{IR}{Information Retrieval}
\newacronym{qa}{QA}{Question Answering}
\newacronym{bnc}{BNC}{British National Corpus}
\newacronym{ukw}{UKWaC}{UK Web Pages}
\newacronym{ai}{AI}{Artificial Intelligence}
\newacronym{gdc}{GDC}{Gothenburg Dialogue Corpus}
\newacronym{dialogpt}{DialoGPT}{Dialogue Generative Pre-trained Transformer}
\newacronym{gpt}{GPT}{Generative Pre-trained Transformer}
\newacronym{multiwoz}{MultiWOZ}{Multi-Domain Wizard-of-Oz}
\newacronym{t5}{T5}{Text-to-Text Transfer Transformer}
\newacronym{bart}{BART}{Bidirectional \& Auto-Regressive Transformer}
\newacronym{xlmr}{XLM-R}{Cross-Lingual Model-RoBERTa}
\newacronym{m2m}{M2M}{Many-to-Many multilingual translation model}
\newacronym{bert}{BERT}{Bidirectional Encoder Representations from Transformers}
\newacronym{roberta}{RoBERTa}{Robustly optimized BERT pretraining Approach}
\newacronym{elmo}{ELMo}{Embeddings from Language Models}
\newacronym{pii}{PII}{personally identifiable information}
\newacronym{qg}{QG}{Question Generation}
\newacronym{tc}{TC}{Text Classification}
\newacronym{pcl}{PCL}{Patronising and Condescending Language}
\newacronym{gus}{GUS}{Genial Understander System}
\newacronym{gmb}{GMB}{Groningen Meaning Bank}
\newacronym{wsd}{WSD}{Word Sense Disambiguation}
\newacronym{ccby4}{CC-BY4}{Creative Commons Attribution 4.0}
\newacronym{ci}{CI}{confidence interval}
\newacronym{bleu}{BLEU}{bilingual evaluation understudy}
\newacronym{gdpr}{GDPR}{General Data Protection Regulation}
\newacronym{svm}{SVM}{support vector machine}
\newacronym{vs}{VS}{vector space}
\newacronym{vsm}{VSM}{vector space model}
\newacronym{nltk}{NLTK}{natural language toolkit}
\newacronym{tf-idf}{tf-idf}{term frequency-inverse document frequency}
\newacronym{pca}{PCA}{Principal Component Analysis}
\newacronym{svd}{SVD}{Singular Value Decomposition}
\newacronym{lsi}{LSI}{Latent Semantic Indexing}
\newacronym{plsi}{PLSI}{Probabilistic Latent Semantic indexing}
\newacronym{lda}{LDA}{Latent Dirichlet Allocation}
\newacronym{lm}{LM}{language model}
\newacronym{bilm}{biLM}{bidirectional language model}
\newacronym{pos}{PoS}{part of speech}
\newacronym{nnlm}{NNLM}{neural network language model}
\newacronym{bpe}{BPE}{byte-pair encoding}
\newacronym{oov}{OOV}{out-of-vocabulary}
\newacronym{imdb}{IMDB}{Internet Movie Database}
\newacronym{lr}{LR}{learning rate}
\newacronym{cus}{CUS}{Credibility unanimous score}
\newacronym{ie}{IE}{Information Extraction}
\newacronym{rl}{RL}{reinforcement learning}
\newacronym{mdl}{MDL}{minimal dependency length}
\newacronym{mlm}{MLM}{masked language model}
\newacronym{rq}{RQ}{research questions}

\title{Vector Representations of Idioms in Conversational Systems}

\name{Tosin Adewumi*, Foteini Liwicki and Marcus Liwicki} 

\address{ML Group,   \\
         EISLAB,\\
         Luleå University of Technology, Sweden\\
         firstname.lastname@ltu.se\\}

\abstract{
We demonstrate, in this study, that an open-domain conversational system trained on idioms or figurative language generates more fitting responses to prompts containing idioms.
Idioms are part of everyday speech in many languages, across many cultures, but they pose a great challenge for many \acrfull{nlp} systems that involve tasks such as \acrfull{ir} and \acrfull{mt}, besides conversational \acrshort{ai}.
We utilize the \acrfull{pie}-English idioms corpus for the two tasks that we investigate: classification and conversation generation.
We achieve \acrfull{sota} result of 98\% macro F1 score on the classification task by using the \acrshort{sota} \acrshort{t5} model.
We experiment with three instances of the \acrshort{sota} dialogue model, \acrfull{dialogpt}, for conversation generation.
Their performances are evaluated using the automatic metric perplexity and human evaluation.
The results show that the model trained on the idiom corpus generates more fitting responses to prompts containing idioms 71.9\% of the time, compared to a similar model not trained on the idioms corpus. 
We contribute the model checkpoint/demo and code on the HuggingFace hub for public access.
 \\ \newline  }

\begin{document}

\maketitleabstract

\section{Introduction}
Open-domain conversational systems struggle to generate fitting responses to prompts containing idioms or figures of speech.
Performance of such systems drop considerably when given context with idioms \cite{jhamtani-etal-2021-investigating}.
This challenge is not limited to open-domain conversational systems alone.
\acrfull{nlp} systems involving tasks such as \acrfull{wsd}, \acrfull{ir}, and \acrfull{mt} also face challenges with regards to idioms \cite{korkontzelos2013semeval,adewumi2021potential}.
The research question we address in this study is "does an open-domain conversational system that is idiom-aware generate more fitting responses to prompts containing idioms?"
In order to investigate this question, we compare three instances of the same \acrshort{sota} model: \acrfull{dialogpt} by \newcite{zhang2020dialogpt}, whereby two are exposed by training to a dedicated idioms dataset and one is not.
We choose the \acrfull{pie}-English idioms corpus by \newcite{adewumi2021potential} for this purpose.
We evaluate the models using the automatic metric, perplexity, and human evaluation in two similar, but different, sets of experiments.

Two separate \acrshort{nlp} tasks are carried out in this study.
The first involves idiom identification or classification and the second involves conversation generation.
Idiom identification can be essential for other \acrshort{nlp} systems.
There are usually two methods to idiom detection: type-based (depends on the expression) and token-based (depends on the context of usage) \cite{peng2015classifying,li2009classifier,sporleder2010idioms}.
In this work, we focus on token-based, the latter.

The key contributions of this work are (1) the demonstration that an open-domain conversational system that is idiom-aware generates diverse and more fitting responses to prompts containing idioms than one that is not and (2) we obtain \acrshort{sota} result in the classification task over the \acrshort{pie}-English idioms corpus by using the \acrshort{sota} \acrfull{t5} base model, compared to the baseline result obtained by \cite{adewumi2021potential}.
The IdiomWOZ model checkpoint and code are hosted on the HuggingFace hub \footnote{huggingface.co/tosin/dialogpt\_mwoz\_idioms}.
Its model card is available in the appendix.
The remaining parts of this paper are as follows.
The Material and Methods section points out the datasets and models used.
It also describes the details of the experiments carried out and the metrics of evaluaton.
The Results section gives results of the experiments for the two tasks, the error analysis and evaluator feedback.
The Related Work section briefly discusses past efforts that are connected to this study.
The Limitation section describes some of the limitation of this work and the Conclusion section summarizes this work.

\section{Materials and Methods}
All the experiments were performed on a shared DGX-1 machine with 8 x 32 Nvidia V100 GPUs.
The operating system on the server is Ubuntu 18.
It has 80 CPU cores.

\subsection{Datasets}

\paragraph{\acrfull{multiwoz} dataset}
MultiWOZ is a large, multi-domain, multi-topic, and multi-task conversational dataset, originally designed for task-oriented dialogues \cite{budzianowski-etal-2018-multiwoz}.
It is a labelled collection of human-human written conversations and consists of more than 10,000 dialogues distributed between 70\% multi-domain
and 30\% single domain dialogues.
The data-acquisition pipeline involved crowd-sourcing without the hiring of professional annotators.
\newcite{budzianowski-etal-2018-multiwoz} considered different dialogue scenarios that includes requests for basic information about attractions through to booking a hotel room, restaurant, train, or taxi between cities.
Additional domains covered are hospital and police.
It has been a standard benchmark for different dialogue problems.
It was used in neural context-to-response generation experiments by \newcite{budzianowski-etal-2018-multiwoz} and adapted for open-domain conversational systems by \newcite{adewumi2021sm,adewumi2022itakuroso}.
There are several versions of the dataset, with each new one bringing improvements \cite{eric-EtAl:2020:LREC}.

\paragraph{\acrshort{pie}-English idioms corpus}
For both the classification and conversation generation tasks investigated in this work, the \acrshort{pie}-English idioms corpus is used.
Hence, we discuss some of the characteristics of the corpus.
It is based on example sentences from two base corpora: the \acrfull{bnc} and \acrfull{ukw}  \cite{ferraresi2008introducing}.
About 76.94\% of the samples are metaphors, making it the largest class in the dataset.
Table \ref{idexamples} shows some examples from the dataset and the short data statement that captures other key characteristics of the dataset, as given by \newcite{adewumi2021potential}, is given afterwards.
Before training in both tasks, the corpus is split in the ratio 80:10:10 for training, dev and test sets, respectively.
Three runs per experiment are conducted in order to calculate the average accuracies, F1 scores, perplexities and standard deviation.
All cases are lowered and any \textit{html} tags removed, if any, as a pre-processing step.
Special characters and numbers are also removed.
Data shuffling is performed before training.

\begin{table*}[h]
\centering
\caption{\label{idexamples} Samples from the \acrshort{pie}-English idioms corpus.}
\begin{tabular}{p{.03\textwidth}p{.67\textwidth}p{.15\textwidth}}
\hline
\textbf{No} & \textbf{Samples} & \textbf{Class}
\\
\hline
1 & 
Carry the day & Metaphor
\\
\hline
2 &  Does the will of the Kuwaiti parliament transcend the will of the Emir and does parliament carry the day? & Metaphor
\\
\hline
3 & 
Time flies & Personification
\\
\hline
4 &  Eighty-four!' she giggled.' How time flies & Personification
\\
\hline
5 & 
 As clear as a bell & Simile
\\
\hline
6 & It sounds as clear as a bell & Simile
\\
\hline
7 & 
 Go belly up & Euphemism
\\
\hline
8 & If several clubs do go belly up, as Adam Pearson predicts. & Euphemism
\\
\hline
9 & 
The back of beyond & Hyperbole
\\
\hline
10 &  There'd be no one about at all in the back of beyond. & Hyperbole
\\
\hline
11 & 
 "Why couldn't you just stay in the back of beyond?" she said. & Hyperbole
\\
\hline
\end{tabular}
\end{table*}

\begin{quote}
    \textbf{\textit{Short data statement for the \acrshort{pie}-English idioms corpus.}}\\
    This is the \acrfull{pie}-English idioms corpus for training and evaluating models in idiom identification.\\
    The licence for using this dataset comes under CC-BY 4.0.\\
    Total samples: 20,174\\
    There are 1,197 total cases of idioms and 10 classes.\\
    Total samples of euphemism (2,384), literal (1,140), metaphor (14,666), personification (448), simile (1,232), parallelism (64), paradox (112), hyperbole (48), oxymoron (48), and irony (32).
\end{quote}

In order to use the \acrshort{pie}-English idioms corpus for the second task of conversation generation as intended, we make the assumption that the corpus is suitable as a conversational dataset of dialogue turns.
This assumption is valid because the sentences of the turns discuss the same cases of idioms despite being drawn from different examples from the base corpora.

\subsection{Classification}

Two \acrshort{sota} pretrained models are compared in the classification task.
These are the \acrfull{bert} \cite{devlin2018bert} and \acrshort{t5} \cite{JMLR:v21:20-074} models from the HuggingFace hub \cite{wolf-etal-2020-transformers}.
Both models involve their base versions.
We use batch size of 64 and 16 for \acrshort{bert} and \acrshort{t5}, respectively.
The total training epochs for both are 6.
Although the choice of hyperparameters can have significant impact on the performance of embeddings or models \cite{adewumi2020word2vec,adewumi2020exploring}, we do not carry out extensive hyperparameter exploration.

\subsection{Conversation Generation}

Three instances of the \acrshort{sota} \acrshort{dialogpt} model are compared in the conversation generation task.
The first instance (IdiomWOZ) is created from the model checkpoint by \newcite{adewumi2021sm}, which has been trained on the MultiWOZ dataset and is available on the HuggingFace hub\footnote{huggingface.co/tosin/dialogpt\_mwoz}.
This is achieved by finetuning on the \acrshort{pie}-English idioms corpus.
The second instance (IdiomOnly) is created from the original \acrshort{dialogpt}-medium model by \newcite{zhang2020dialogpt} by finetuning on the same idioms corpus.
The model checkpoint by \newcite{adewumi2021sm} for the first instance is also based on the medium version of \acrshort{dialogpt}.
The third instance (MultiWOZ) is the model checkpoint by \newcite{adewumi2021sm}, which was trained on the MultiWOZ dataset.
For all the three instances, we set the decoding algorithm as top-k (k=100) and top-p (p=0.7).
Other hyperparameters are maximum decoding length of 200 tokens, temperature of 0.8, and maximum ngram repeat limit of 3.

All three model checkpoints are then used to generate three transcripts of conversations in a first set of experiments.
Ninety-four random numbers were generated and used to select the same prompts from the two test sets: the \acrshort{multiwoz} and the \acrshort{pie}-English idioms corpus.
The prompts are fed to the three models.
For each dataset, thirty-two of the prompts are for generation and fifteen prompts with their test set responses (for credibility) are selected.
In the second set of experiments, which has the objective of finding out fitting and diverse responses to idiom prompts, sixty-two random numbers were generated.
Thirty-two are from the idioms test set and used as prompts for both the IdiomWOZ and MultiWOZ, while the remaining thirty are credibility conversations from the \acrshort{multiwoz} test set.
The credibility conversations are to determine the suitability of the evaluators, hence the responses to these prompts are the corresponding responses from the test sets.
They are distributed at regular intervals within the transcripts.

\subsubsection{Evaluation}
Automatic metrics, such as \acrshort{bleu} or ROUGE \cite{lin-2004-rouge,papineni2002bleu}, which are common for \acrfull{nlg} tasks like \acrshort{mt} \cite{gehrmann-etal-2021-gem,vaswani2017attention}, are sometimes viewed as inadequate for conversational systems \cite{jurafsky2020speech,liu2016not}.
This is because they do not correlate well with human assessment \cite{reiter201020}.
In this work, we use another common metric, perplexity, which is also used by \newcite{adiwardana2020towards}.
Smaller perplexity values show that a model fits the data better as it measures how well a probability model predicts a sample, thereby corresponding to the effective size of the vocabulary \cite{aggarwal2012survey}.

For human evaluation, evaluators (or annotators) were recruited on Slack\footnote{slack.com}.
They are second/L2 (but dominant) speakers of English and are unbiased respondents who did not take part in the training of the models.
For the evaluation of the transcripts, \textit{Instruction 1} and \textit{Instruction 2} below are the instructions for transcripts from the first and second set of experiments, respectively.
Three valid evaluated transcripts from three annotators are accepted per set of experiments.
The first set of transcripts are evaluated for human-likeness while the second set are based on two characteristics: more fitting and more diverse responses.

\begin{quote}
    \textbf{\textit{Instruction 1:}}
    Here are 94 different conversations by 2 speakers. Please, write Human-like (H) or Non-human-like (N) or Uncertain (U), based on your own understanding of what is human-like. Sometimes the speakers use idioms. If you wish, you may use a dictionary.
\end{quote}

\begin{quote}
    \textbf{\textit{Instruction 2:}}
    Person 2 \& Person 3 respond to Person 1. Please, write which (2 or 3) is the a) more fitting response \& b) more diverse response (showing variety in language use).
\end{quote}

\subsubsection{Credibility Unanimous Score (\acrshort{cus})}
\label{credibility}
In order to measure \acrfull{iaa} of the conversation transcripts, we use \acrshort{cus}, which was introduced by \newcite{adewumi2022itakuroso}.
It is more intuitive, easier to calculate (based on percentages) and appears less sensitive to changes in the number of categories being evaluated, when compared to Flies Kappa (\textit{k}).
Fleiss Kappa (\textit{k}) is known to be restrictive in its interpretation, depending on the number of categories \cite{landis1977measurement}, as Kappa is lower when the categories are more \cite{sim2005kappa}.
According to \newcite{adewumi2022itakuroso}, the assumption behind \acrshort{cus} is that if homogeneous samples may be used for checking the credibility of the annotators, then they may be used for establishing their agreement over the transcript.
The agreement is based on unanimous votes on the homogeneous samples that are introduced.
These samples may be viewed as a significant subset of the entire transcript, particularly when there's a minimum of 30 samples, thereby fulfilling the central limit theorem.
The probability of obtaining high \acrshort{cus} rises when the benchmark score for annotator credibility is high.

\section{Results}

\begin{table*}[h!]
\centering
\caption{\label{res1}Average accuracy \& F1 results
\footnotesize{(sd - standard deviation) *data split ratio 85:15 for training:dev sets.}}
\begin{tabular}{c|c|c|c|c|c|c}
\textbf{Model} &
\multicolumn{2}{c|}{\textbf{Accuracy}} &
\multicolumn{2}{c|}{\textbf{weighted F1}} &
\multicolumn{2}{c}{\textbf{macro F1}}
\\
\hline
 & dev (sd) & test (sd) & dev (sd) & test (sd) & dev (sd) & test (sd)
\\
\hline
BERT & 0.96 (0) & 0.96 (0) & 0.96 (0) & 0.96 (0) & 0.75 (0.04) & 0.73 (0.01)
\\
\hline
T5 & 0.99 (0) & 0.98 (0) & 0.98 (0) & 0.98 (0) & 0.97 (0) & 0.98 (0)
\\
\hline
BERT* \cite{adewumi2021potential} & 0.93 & - & 0.95 & - & - & -
\\
\hline
\end{tabular}
\end{table*}

\begin{figure*}[h!]
\centering
\includegraphics[width=16cm]{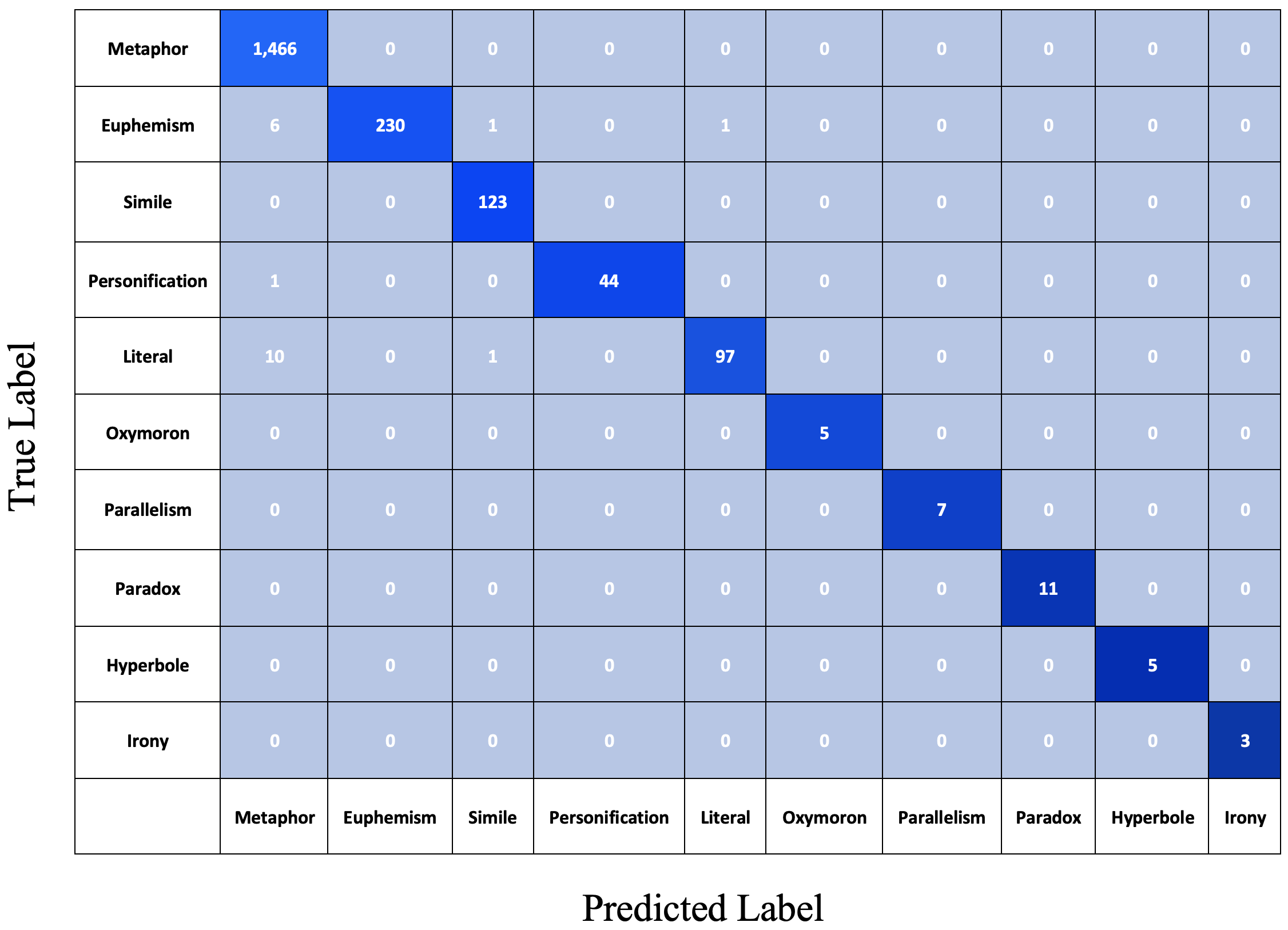}
\caption{Confusion matrix for \acrshort{t5} model on the \acrshort{pie}-English idioms corpus test set. \label{fig1}}
\end{figure*}

\subsection{Classification}
Table~\ref{res1} shows that the \acrshort{t5} model outperforms the \acrshort{bert} model.
It also outperforms the best model from \newcite{adewumi2021potential}, another \acrshort{bert} implementation.
Results from \newcite{adewumi2021potential} do not provide standard deviation values and they report results only on the training and dev sets split of 85:15 ratio.
From the results, it appears that the \acrshort{pie}-English idioms corpus is not overly challenging, at least for the \acrshort{t5} model, because of the high scores obtained.
This may be due to the fact that the length of each sample is one sentence, or at most 2 sentences, in the corpus.
The results are statistically significant as the p-value (p $<$ 0.0001) of the two-sample t-test for the difference of two means (of the macro F1 scores) is smaller than alpha (0.05).

\paragraph{Error Analysis}

Figure~\ref{fig1} shows the confusion matrix of the results of the \acrshort{t5} model for the predictions against the true labels for the test set of the idioms corpus.
The model performs substantially well even for classes that have few samples in the training set, such as \textit{hyperbole} and \textit{irony}.
It, however, struggles mostly in correctly classifying the \textit{literals}.
It misclassified about 9.3\% of them as \textit{metaphor}, presumably because it is the largest class in the dataset.

\subsection{Conversation generation}

We can observe from Table~\ref{pepid} that the \acrshort{multiwoz} model from \newcite{adewumi2021sm} has the lowest average perplexity, when compared with the other two new models.
This is likely because the \acrshort{multiwoz} data the model was trained on is larger than the idioms corpus.
The p-value (p $<$ 0.0001) of the two-sample t-test for the difference of two means (for the IdiomWOZ and IdiomOnly) is smaller than alpha (0.05), hence the results are also statistically significant.
Despite the average perplexity for the IdiomOnly model being lower than the IdiomWOZ, we chose to generate responses and conduct human evaluation on the latter.
This is because one of its runs had a lower perplexity, which may be deduced from the standard deviation.
Besides, perplexity alone may not be sufficient to tell how good a model is \cite{roller-etal-2021-recipes,hashimoto-etal-2019-unifying}.

\begin{table*}[h!]
\caption{\label{pepid}Average perplexity results. \footnotesize{(sd - standard deviation)}
}
\centering
\begin{tabular}{c|c|c}
\textbf{Model} &
\multicolumn{2}{c}{\textbf{Perplexity}}
\\
\hline
 & dev (sd) & test (sd)
\\
\hline
IdiomWOZ & 201.10 (34.82) & 200.68 (34.83)
\\
\hline
IdiomOnly & 189.92 (1.83) & 185.62 (2.05)
\\
\hline
MultiWOZ \cite{adewumi2021sm} & 6.41 (-) & 6.21 (-)
\\
\hline
\end{tabular}
\end{table*}

\begin{table*}[h!]
\caption{\label{humeval}Human evaluation results of 3 annotators on 3 classes for 64 single-turn conversations.
}
\centering
\begin{tabular}{lccccc}
\hline
\textbf{Model} &
\multicolumn{4}{c}{\textbf{Scale (majority votes)}} & \textbf{\acrshort{cus}}
\\
 &  H (\%) & U (\%) & N (\%) & 3-way (\%) & \%
\\
\hline
IdiomWOZ & 39.1 & 10.9 & 37.5 & 12.5 & 80
\\
IdiomOnly & 15.6 & 12.5 & 60.9 & 10.9 & 80
\\ 
MultiWOZ & 62.5 & 1.6 & 32.8 & 3.1 & 80
\\
\hline
 &
\multicolumn{4}{c}{\textbf{unanimous votes - 3/3}} &
\\
\hline
IdiomWOZ & 20.3 & 0 & 12.5 & - & 80
\\
IdiomOnly & 6.3 & 0 & 31.3 & - & 80
\\ 
MultiWOZ & 45.3 & 0 & 23.4 & - & 80
\\
\hline
 &
\multicolumn{4}{c}{\textbf{idioms only maj. votes (32 samples)}} &
\\
\hline
IdiomWOZ & 30 & 23.3 & 33.3 & 13.3 & 80
\\
IdiomOnly & 26.7 & 0.2 & 36.7 & 16.7 & 80
\\ 
MultiWOZ & 26.7 & 3.3 & 66.7 & 3.3 & 80
\\
\hline
\end{tabular}
\end{table*}

\begin{table*}[h!]
\caption{\label{humeval2}Human evaluation results of 3 annotators on 3 classes for 32 single-turn conversations.
}
\centering
\begin{tabular}{lc|cc}
\hline
\textbf{Model} &
\multicolumn{2}{c}{\textbf{Scale (majority voting)}} & \textbf{\acrshort{cus}} 
\\
 &  More fitting (\%) & More diverse (\%) & \%
\\
\hline
IdiomWOZ & 71.9 & 28.1 & 80 
\\ 
MultiWOZ & 28.1 & 71.9 & 80 
\\
\hline
 &
\multicolumn{2}{c}{\textbf{unanimous votes - 3/3}}  &
\\
\hline
IdiomWOZ & 46.9 & 9.4 & 80 
\\ 
MultiWOZ & 9.4 & 15.6 & 80 
\\
\hline
\end{tabular}
\end{table*}

Human evaluation results are presented in Tables~\ref{humeval} and ~\ref{humeval2}.
The former is based on transcripts of 64 single-turn conversations for the first set of experiments and the latter is based on 32 single-turn conversations for the second set of experiments, after the removal of the 30 credibility conversations from each.
We observe, from Table~\ref{humeval}, that both the MultiWOZ and IdiomWOZ models have more human-like single-turn conversations than other categories, under majority votes.
The MultiWOZ model has the most human-like conversations.
However, considering idioms only prompts, the IdiomWOZ model has the most human-like conversations.
From Table~\ref{humeval2}, IdiomWOZ has more fitting conversations than the MultiWOZ, though the opposite is the case with regards to more diverse conversations.
This could be due to the evaluators' interpretation of what is diverse.
For all the evaluations, we observe that there is \acrshort{cus} of 80\%.
The \acrshort{cus} is the same across sections in each table since the same transcript is involved for each section.

\section{Discussion \& Evaluator Feedback}

Tables~\ref{transcript} and ~\ref{transcript2} show some single-turn conversations from the second transcript.
Person 1 is the prompt from the \acrshort{pie}-English idioms test set.
Person 2 is the IdiomWOZ model and it responds more with idioms.
Person 3 is the MultiWOZ model.

\begin{table*}[h!]
\centering
\caption{Excerpt from the second transcript of single-turn conversations. Person 2 is the IdiomWOZ\label{transcript}}
\begin{tabular}{p{.48\textwidth}p{.48\textwidth}}
\hline
\textbf{Person 1} & \textbf{Person 2}
\\
\hline
 He knew his father's grievance ; and put off the evil day as long as he possibly could & 
he's dead meat
\\
\hline
achieving results that fly in the face of realism & I'm not here I'm just a fly on the wall
\\
\hline
 does that mean Jane is off the hook ? &  She had the nerve to call time on her three-year spell in charge
\\
\hline
he's still sulking ; sounding like a broken record & the whole caboodle
\\
\hline
 we need new ways to do this ; to chew the fat ; to check ourselves out & the Government's approach to the environment is geared more towards addressing the environment rather than helping the environment
\\
\hline
 the 24-year-old striker is self-critical enough to insist he is not yet firing on all cylinders &  You're the one up to your neck in it
\\
\hline
 remind her to take it off the hook & 
the bookie took it on the nod
\\
\hline
A lengthy break is in order ; so make sure you keep your nose clean &  they are under a cloud because of fears about US President's wife Hillary Clinton
\\
\hline
 Utd have shown they are in a league of their own & It's a long shot him signing for Leeds. They're a hard nut to crack
\\
\hline
 from now until the year dot &  he was the apple of her eye
\\
\hline
it's a good idea to go the distance &  it 'll just collapse and give up the ghost somewhere in South Africa
\\
\hline
\end{tabular}
\end{table*}

\begin{table*}[h!]
\centering
\caption{Excerpt from the second transcript of single-turn conversations. Person 3 is the MultiWOZ\label{transcript2}}
\begin{tabular}{p{.48\textwidth}p{.48\textwidth}}
\hline
\textbf{Person 1} & \textbf{Person 3}
\\
\hline
 He knew his father's grievance ; and put off the evil day as long as he possibly could & 
That's great; thanks.
\\
\hline
 achieving results that fly in the face of realism & Yes; realism. I like it.
\\
\hline
does that mean Jane is off the hook ? &  I'm looking for a place to go in the centre.
\\
\hline
he's still sulking ; sounding like a broken record & Well then I suppose I need to find something else to do for you.
\\
\hline
 we need new ways to do this ; to chew the fat ; to check ourselves out & To each their own. Let me know if you need any more info.
\\
\hline
the 24-year-old striker is self-critical enough to insist he is not yet firing on all cylinders &  Well if you'll excuse me... let me ask my wife. Does she think we should try for a little earlier; say 20:30?
\\
\hline
 remind her to take it off the hook & I'm sorry; I forgot to hang up. Good bye.
\\
\hline
A lengthy break is in order ; so make sure you keep your nose clean & I am looking for a restaurant in the centre of town that serves unusual food.
\\
\hline
 Utd have shown they are in a league of their own & I'm sorry; but I'm not sure what you're looking for. Could you please restate your requirements?
\\
\hline
from now until the year dot &  I'm sorry; but I don't understand your request. Would you like to try a different restaurant?
\\
\hline
it's a good idea to go the distance & I can help with that! What are your departure and destination sites?
\\
\hline
\end{tabular}
\end{table*}

The IdiomWOZ model, whose responses are deemed as more fitting, appears to confirm previous study, which shows that metaphors have more emotional impact than their literal equivalent \cite{mohammad-etal-2016-metaphor}.
Idioms usually make utterances more colorful (or rich) and diverse.
Hence, simply reducing idioms to their literal form before feeding \acrshort{nlp} models as practised by \newcite{jhamtani-etal-2021-investigating} may not adequately address the challenge since it implies the models still are incapable of understanding the idioms and because some idioms have more than one literal expression.

Feedback from some of the evaluators suggest the use of idioms complicated the evaluation task a bit but it was sometimes useful in identifying which of the two conversations was a more fitting or diverse option.
They found it relatively easier evaluating the human-likeness characteristic in the first set of experiments but had to resort to using a dictionary for the idioms in certain cases.
Some of the conversations were marked non-human-like because there was lack of connection between the prompt and the response.
In the second set of transcripts, some conversations are evaluated as more fitting when the responses answer the prompts directly while some are evaluated as more diverse when the response is not a direct answer but elicits further discussion.

\section{Related Work}
\newcite{jhamtani-etal-2021-investigating} observed that performance dipped when some deep models were evaluated on two open-domain dialogue datasets: DailyDialog and PersonaChat, with regards to figurative language \cite{li-etal-2017-dailydialog,zhang-etal-2018-personalizing}.
They compared \acrfull{gpt}-2 to four other models over the datasets and noticed the drop in performance among most models.
In their work, however, they proposed transforming figurative language (including idioms) to their literal form before feeding the models.
Idiom detection usually takes two approaches: type-based and token-based \cite{peng2015classifying,cook2007pulling,li2009classifier,sporleder2010idioms}.
The type-based approach attempts to determine if an expression is an idiom, perhaps through automatic compilation of an idiom list from a corpus \cite{sporleder-etal-2010-idioms}.
The token-based approach relies on context for distinguishing idioms \cite{korkontzelos2013semeval,sporleder2010idioms}.
Non-contextual word embeddings (such as word2vec) are used for identifying metaphors \cite{mao-etal-2018-word}.
However, such an approach may underperform \cite{mao-etal-2018-word}.
\newcite{peng-etal-2015-classifying} use word2vec to obtain vectors from text8 corpus.
Their algorithm, based on the assumption that literal vectors are distinguished from idiom vectors by the larger inner product they produce, uses inner product of context word vectors with vector representing target expression.
\newcite{bizzoni-etal-2017-deep} use word2vec and an \acrshort{ann} with 1 hidden layer for detecting metaphors.
\newcite{diab2009verb} perform binary classification using \acrfull{svm}, producing literal and idiomatic expressions on a subset of the VNC-Token.

\section{Limitation}
The \acrshort{pie}-English idioms corpus that is used in this work, though relatively large and probably the first with as much as 10 classes, does not represent all the possible classes of idioms available.
This may affect the performance of the \acrshort{ml} models.
Furthermore, there may be risks of offensive language or biased outputs from the model checkpoints, since data used for pretraining the deep models are from public online sources that are known to contain such risks \cite{zhang2020dialogpt}.

\section{Conclusions}

We answer the research question "does an open-domain conversational system that is idiom-aware generate more fitting responses to prompts containing idioms?"
The answer is yes.
It is, therefore, important to train open-domain conversational systems on idioms data, so as to achieve diversity and more fitting responses in \acrshort{ml} models.
This is especially since idioms are part of everyday speech in many cultures \cite{alm2003figures}.
We achieve \acrshort{sota} classification results on the \acrshort{pie}-English idioms corpus, using the \acrshort{t5} model.
Future efforts may be directed at exploring more datasets of figurative language or idioms and more diverse \acrshort{sota} models for training.

\onecolumn
\newpage

\appendix
\section*{Appendix}

\begin{table*}[h]
\centering
\begin{tabular}{p{.15\textwidth}|p{.8\textwidth}}
\multicolumn{2}{l}{Model card for DialoGPT(medium)-MultiWOZ, trained on the \acrshort{pie}-English idioms corpus}
\\
 \hline
\textbf{} &
\textbf{Details}
\\
\hline
Model rationale & This is a fine-tuned model of DialoGPT (medium)-MultiWOZ v2.2 on the \acrfull{pie}-English idioms corpus.
It is intended to be used as an idiom-aware conversational system.
\\
\hline
Limitation & The dataset it's trained on is limited in scope, as it covers only 10 classes including metaphor, simile, euphemism, parallelism, personification, oxymoron, paradox, hyperbole, irony and literal.
\\
\hline
Result & The perplexity achieved on the test and validation sets are 160.9 and 160.46, respectively.
\\
\hline
 & \textbf{Others}
\\
\hline
Model link & https://huggingface.co/tosin/dialogpt\_mwoz\_idioms
\\
\hline
More & Information about the original pre-trained model can be found at https://huggingface.co/tosin/dialogpt\_mwoz
\\
\hline
Licence & CC-BY 4.0.
\\
\hline
 & \textbf{Generation example from an interactive environment:}
\\
\hline
\textbf{Role} & \textbf{Response}
\\
\hline
User & remind her to take it off the hook.
\\
\hline
Bot & the bookie took it on the nod
\\
\hline
User & A lengthy break is in order ; so make sure you keep your nose clean
\\
\hline
Bot & they are under a cloud because of fears about US President's wife Hillary Clinton
\\
\hline
User & Some families breed like rabbits and send us all their male heirs
\\
\hline
Bot & he was out for the count
\\
\hline
\end{tabular}
\caption{\label{appsvanalogy}}
\end{table*}

\twocolumn
\section{Bibliographical References}\label{reference}

\bibliographystyle{lrec2022-bib}
\bibliography{lrec2022-example}


\printglossary[type=\acronymtype]

\end{document}